\documentclass{article}
\usepackage{spconf,amsmath,amssymb,graphicx,booktabs,tabularx,array,float,hyperref,xcolor,multirow,microtype}

\newcommand{\cmark}{\ensuremath{\checkmark}}

\newcommand{\datasetname}{\textsc{VietPrism}}
\newcommand{\ours}{\textbf{\datasetname}}
\let\standardthebibliography\thebibliography
\renewcommand{\thebibliography}[1]{\standardthebibliography{#1}\small\hbadness=3000\relax}
\newif\ifshowtablecitations
\showtablecitationsfalse
\title{{\sc \datasetname{}}: A Large-Scale Vietnamese Speech and Deepfake Corpus with Diverse Dialects and Code-Switching}
\name{Minh Hoang$^{\dagger}$, Thai Le$^{\ddagger}$}
\address{Independent Researcher$^{\dagger}$\;\; Indiana University, Bloomington, USA$^{\ddagger}$}
\begin{document}
\raggedbottom
%
\maketitle
\begin{abstract}
Vietnamese speech research is constrained by resources that isolate automatic
speech recognition from speaker, dialect, code-switching, and deepfake
analysis. We introduce \datasetname, an open, multi-domain corpus that brings
these dimensions together at scale: 993.4 hours and 403,941 bona fide
utterances from 1,262 verified speakers across 8,388 real-world videos. To our
knowledge, it is the first large-scale Vietnamese corpus to jointly provide
transcripts, consistent speaker identities, five dialect groups, and naturally
occurring Vietnamese--English code-switching, which constitutes nearly half of
the corpus by duration. We further create over 3.1K hours of spoof speech with four
open-source and commercial synthesis systems. Every spoof is conditioned on a
verified speaker reference and paired with a transcript- and speaker-matched
bona fide utterance, enabling unique controlled evaluation with reduced lexical and
identity confounds. Zero-shot evaluation of five pretrained multilingual
detectors reveals striking brittleness: EER greatly varies across
detector--generator pairings, while recent multilingual detector DFA-1B degrades from 16.3\% to 33.6\% as
speaker similarity increases. Dialect-stratified results expose further
model-dependent disparities. By unifying natural linguistic diversity with
controlled spoof generation, \datasetname{} provides a challenging foundation
for Vietnamese speech modeling and trustworthy audio-deepfake detection.

\end{abstract}
\section{Introduction}
\label{sec:intro}

\begingroup
\ifshowtablecitations\else
\renewcommand{\cite}[1]{}
\fi
\begin{table}[tb!]
\centering

\label{tab:unified-datasets-compact}
\footnotesize
\setlength{\tabcolsep}{1.5pt}
\renewcommand{\arraystretch}{0.96}
\begin{tabular}{@{}l c c c c c c r@{}}
\toprule
Dataset & CS & Dial. & Spk. B/S & Hrs. B/S & Avg. B/S & Tr & Utt. B/S \\
\midrule
\multicolumn{8}{@{}l}{\textit{Speech corpora}} \\
VNSC~\cite{le2004spoken} & - & 3 & 50 & 100 & - & - & - \\
VN-LVCSR~\cite{thang2009vietnamese} & - & - & - & 25 & - & - & - \\
VIVOS~\cite{luong2016non} & - & - & 65 & 15 & 4.5 & \cmark & 12K \\
VDSPEC~\cite{hung2016statistical} & - & 3 & 150 & 45.12 & 10 & - & - \\
Viettel-CC~\cite{nguyen2017development} & - & - & - & 85.8 & - & - & - \\
Do-S1~\cite{do2018development} & - & - & - & 6 & - & - & - \\
Do-S2~\cite{do2018development} & - & - & - & 6.5 & - & - & - \\
Do-L~\cite{do2018development} & - & - & - & 900 & - & - & - \\
FOSD~\cite{tran2020fosd} & - & - & - & 30 & 4.2 & \cmark & 26K \\
CV~\cite{ardila2020common} & - & - & 417 & 23 & 4.0 & \cmark & 21K \\
CanVEC~\cite{nguyen2020canvec} & {\cmark} & - & 45 & 10 & - & - & - \\
VinBD~\cite{vinbigdata2020} & - & - & - & 101 & 6.5 & \cmark & 56K \\
VLSP21~\cite{vlsp2021asr} & - & - & - & 280 & - & - & - \\
FLEURS~\cite{conneau2023fleurs} & - & - & - & 13 & 11.4 & \cmark & 4.2K \\
VN-Celeb~\cite{pham2023vietnamceleb} & - & 3 & 1K & 187 & 7.7 & - & 87K \\
ViASR~\cite{nguyen2023viasr} & - & 3 & - & 32 & 25.5 & - & 4.3K \\
VLSP23~\cite{vlsp2023asr} & - & - & - & - & - & - & - \\
Bud500~\cite{Bud500} & - & 3 & - & 511 & 2.8 & \cmark & 649K \\
LSVSC~\cite{tran2024lsvsc} & - & 2+H+M & - & 100 & 6.4 & \cmark & 57K \\
VietMed$\_L$~\cite{leduc2024vietmed} & - & 3+H & 61 & 2K & 6.2 & \cmark & 9.2K \\
VietSpeech~\cite{VietSpeech} & - & 3 & - & 1K & 4.0 & \cmark & 1.03M \\
ViMD~\cite{nguyen2024vimd} & - & 3 & 13K & 103 & 19.5 & \cmark & 19K \\
viVoice~\cite{gia2024vivoice} & - & - & - & 1K & 4.1 & \cmark & 888K \\
VoxVN~\cite{vu2024voxvietnam} & - & - & 1,4K & 261 & 5.0 & - & 188K \\
GS2~\cite{yang2025gigaspeech} & - & - & - & 6K & 4.4 & \cmark & 5.01M \\
PhoAudio~\cite{vu2025zero} & - & - & 735 & 941 & 11.7 & \cmark & 291K \\
OpenBible~\cite{guzman2026openbibletts} & - & - & - & 72 & 8.5 & \cmark & 30K \\
ViMedCSS~\cite{vimedcss} & {\cmark} & - & - & 33 & 7.4 & \cmark & 16K \\
VietSuper~\cite{do2026vietsuperspeech} & - & - & - & 267 & 12.0 & \cmark & 52K \\
\midrule
\multicolumn{8}{@{}l}{\textit{Deepfake datasets}} \\
MLAAD~\cite{muller2024mlaad} & - & - & --/- & --/10.1 & --/7.2 & MT & --/5K \\
VSASV~\cite{hoang2024vsasv} & - & - & 1,141/163 & 199/153 & 7.3/4.5 & - & 98K/123K \\
JMAD~\cite{mawalim2025multilingual} & - & - & -/- & -/- & -/- & - & 5K/0.96K \\
SF-MD~\cite{huang2025speechfake} & - & - & --/2 & --/15.22 & --/2.9 & - & --/19K \\
SEA-Spoof~\cite{wu2025sea} & - & - & -/- & 30.5/49.1 & 4.3/3.2 & \cmark & 26K/55K \\
\midrule
\ours & {\cmark} & 3+O
& \textbf{1,262} & \textbf{993}
& \textbf{8.9/8.9} & \cmark & \textbf{404K} \\
 &  & 
& \textbf{/1262} & \textbf{/3,190}
& &  & \textbf{/1.29M} \\
\bottomrule
\end{tabular}%
\vspace{1pt}
\begin{minipage}{\columnwidth}
\footnotesize
B/S denotes bona fide/spoof; speech corpora contain bona fide audio only.
H/M/O: highland/minority/overseas. --: no audio of that class; -{}: unavailable, unreported, or closed; 3: North, Central, South.
\end{minipage}
\caption{Comparison of \datasetname{} with existing Vietnamese speech and audio-deepfake datasets.}
\end{table}

\endgroup

\begin{figure*}[t]
\centering
\includegraphics[width=\textwidth]{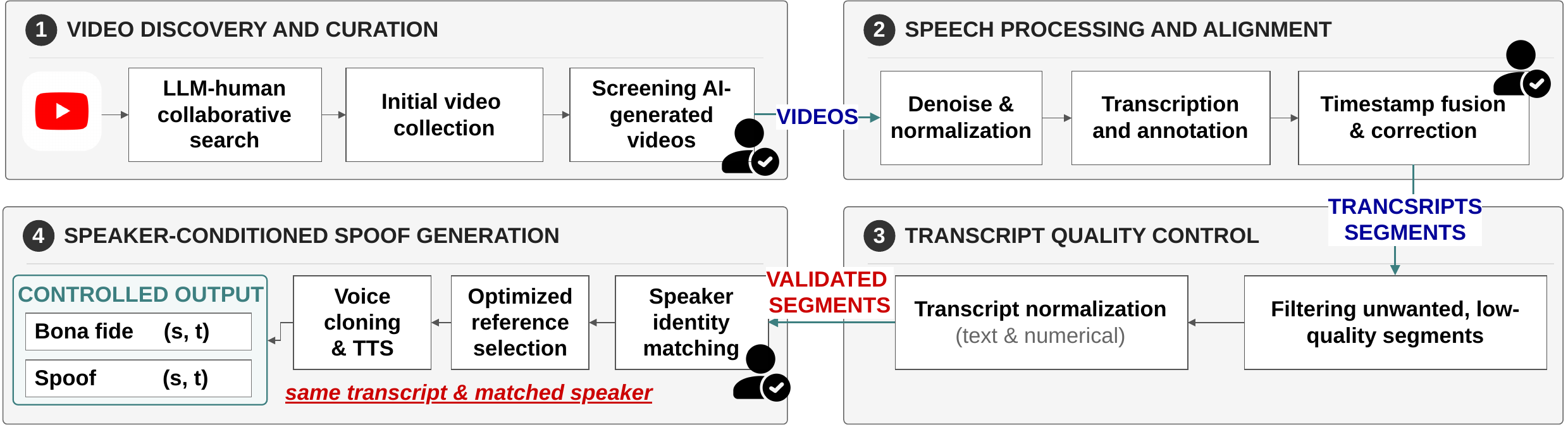}
\caption{Overview of the \datasetname{} data generation pipeline. Human icons depict steps where we involve human screening.}
\label{fig:data-pipeline}
\vspace{-10pt}
\end{figure*}

Vietnamese speech technology has advanced rapidly, supported by corpora such as
VIVOS~\cite{luong2016non}, FPT Open Speech, Common
Voice~\cite{ardila2020common}, Bud500~\cite{pham2024bud500},
VietSpeech~\cite{VietSpeech}, and GigaSpeech~2~\cite{yang2025gigaspeech}.
However, most existing resources were developed primarily for automatic speech
recognition (ASR). Their design consequently emphasizes audio--transcript pairs
while omitting metadata needed beyond ASR. For example, FPT Open Speech,
VinBigData, Bud500, VietSpeech, and GigaSpeech-2 do not provide consistent speaker
identifiers \cite{guzman2026openbibletts,pham2024bud500,VietSpeech,yang2025gigaspeech}.
Conversely, speaker-oriented resources such as Vietnam-Celeb and VoxVietnam do
not provide transcripts, and their released identity labels require additional
care \cite{pham2023vietnam,vu2025voxvietnam}. This gap makes it
difficult to connect utterances from the same person to support tasks such as 
speaker-aware training, voice cloning, speaker verification, and the emerging spoof-detection research within a common resource.

Moreover, there is a distinct limitation concerning linguistic and dialect coverage. Vietnamese
usage in real-world settings is not confined to carefully read,
monolingual sentences. Due to globalization, younger domestic speakers, especially those living overseas, frequently switch between Vietnamese and
English within the same conversation. Such code-switching reflects identity,
community, and communicative context across regional dialects, yet it remains poorly represented in
general-purpose Vietnamese corpora. Existing efforts such as CanVEC~\cite{nguyen2020canvec} captures
natural mixed speech, but contains only 10 hours from 45 bilingual speakers in
Canberra and was developed primarily for sociolinguistic study. Similarly, 
ViMedCSS~\cite{nguyen2026vimedcss} dataset is also restricted to
the medical domain and does not provide speaker identities. These resources
leave a broader gap: to our knowledge, \textit{no open, large-scale, multi-domain
Vietnamese dataset jointly provides naturally occurring Vietnamese--English
code-switching, transcripts, dialect information, and consistent speaker
identities across diverse real-world recordings.}


Furthermore, existing dataset quality presents another gap. Lau et al.~\cite{lau2025data} show substantial speaker imbalance, uneven topic-domain coverage, and limited quality-control documentation in the English Common Voice corpus. Although their quantitative analysis should not be transferred directly to Vietnamese,
the same quality dimensions expose important gaps in the Vietnamese Common
Voice release~\cite{ardila2020common}. The publicly available statistics describe only
22.96 hours and 417 reported contributors, without dialect annotations, topic distribution, or an end-to-end quality audit. Its effective speaker and domain diversity therefore cannot be inferred from duration alone.

To overcome these gaps, we introduce \datasetname, a dataset designed to bridge these gaps and support multiple speech tasks rather than a single benchmark. Our contributions are
threefold. First, we release an open corpus of 993.4 hours and 403,941 utterances from
1,262 speakers, collected from 8,388 real-world videos, making it one of the
largest richly annotated Vietnamese speech corpora
(Table~\ref{tab:unified-datasets-compact}). Second, to our knowledge, \datasetname{} is the first large-scale, open, multi-domain Vietnamese--English code-switching speech
corpus that captures natural usage across diverse real-world topics while
jointly providing transcripts, dialect annotations, and consistent speaker
identities. Third, its speaker-linked annotations and natural 8.9\,s
utterances enable the same resource to support ASR, speaker verification,
dialect and code-switching modeling, voice cloning, and controlled
audio-deepfake detection tasks.

\section{\datasetname{} Dataset}
\label{sec:dataset}

\subsection{Dataset Curation Pipeline}

Figure~\ref{fig:data-pipeline} summarizes our proposed data pipeline.

\noindent \textbf{\underline{Step 1.} Video Discovery and Curation.} We first collect publicly accessible YouTube videos in available highest-quality formats. We manually seed dialectal and Vietnamese--English code-switching
videos and expand collection by channel, then utilize an LLM to iteratively generates
bilingual queries from seed and retrieved videos of similar titles. We cap each channel at three
hours to balance between quantity and diversity, and reject suspected AI-generated videos.

\noindent \textbf{\underline{Step 2.} Speech Processing and Alignment.} Extracted audio is then decoded to WAV, resampled to 16~kHz mono, peak-normalized to 0~dBFS,
loudness-normalized to $-14$~LUFS, and denoised with ClearerVoice-Studio, which shows competitive  performance over other methods such as DeepFilterNet2 in our manual examination
\cite{zhao2025clearervoice,schroter2022deepfilternet2}. We then adopt LLM Gemini 2.5 Pro to process
the enhanced audio for initial transcription, segmentation, diarization, and metadata extraction, retaining attributes, and at the same time maintaining natural fillers. Resulting transcripts' timestamps are then fused
with traditional Montreal Forced Aligner (MFA)~\cite{mcauliffe2017montreal} method, and disagreements exceeding 2 seconds (around 2.4\% of total segments) are manually corrected.

\noindent \textbf{\underline{Step 3.} Transcript Quality Control.} Resulting validated timestamps then help define utterance-level segments. Since we only focus on extracting spoken transcript of one speaker per segment, we then utilize a state-of-the-art LLM Qwen3-Omni-30B-A3B~\cite{xu2025qwen3} with thinking mode to filters unusable, multi-speaker, singing, and mixed samples.

\noindent \textbf{\underline{Step 4.} Speaker-Conditioned Spoof Generation.} To ensure our datasets are also applicable to emerging deepfake research in Vietnamese language, it is important to identify unique speakers appearing across different videos. Thus, all of the collected speaker identifications are manually clustered and verified before extracting 9--9.5-second references for deepfake synthesis using recent open-source and popular commercial deepfake and voice-cloning models including OmniVoice, Higgs Audio v3, VoxCPM2, and MiniMax.
Commercial MiniMax uses primarily Speech 2.8 HD and a smaller Speech 2.6 HD subset, with volume, pitch, and speed are varied for diversity. Each system synthesizes a bona-fide transcript from a verified speaker reference, and the output is paired with its speaker-matched bona-fide segment to control lexical and identity confounds. 

\noindent\textbf{{Human Annotators:}} We recruit a total of 9 adult, native, local Vietnamese speakers of diverse education background from social media, and provide them with a brief instruction on the annotation and screening tasks with averaged payment meeting the local standard for per-hour minimal wage.

\setlength{\tabcolsep}{2pt}
\begin{table}[tb!]
\centering
\footnotesize
\renewcommand{\arraystretch}{1.12}
\begin{tabular}{@{}c@{\hspace{4pt}}p{0.46\columnwidth}rrr@{}}
\toprule
 & \textbf{Sub-category} & \textbf{Percentage} & \textbf{\# Utt.} & \textbf{\# Hours} \\
\midrule
\multirow{5}{*}{\rotatebox[origin=c]{90}{Dialect}}
 & South & 46.7\% & 175,187 & 463.6 \\
 & North & 50.0\% & 215,266 & 496.7 \\
 & Central & 2.0\% & 8,877 & 19.8 \\
 & Southwest & 0.7\% & 4,431 & 7.2 \\
 & Overseas Vietnamese & 0.6\% & 2,172 & 5.9 \\
\midrule
\multirow{9}{*}{\rotatebox[origin=c]{90}{Topic}}
 & Real-estate & 19.8\% & 65,093 & 196.3 \\
 & Beauty, fitness, fashion, and sports & 2.4\% & 10,049 & 23.9 \\
 & Daily life & 13.2\% & 62,163 & 131.1 \\
 & Entertainment & 10.8\% & 54,424 & 107.7 \\
 & Knowledge & 38\% & 151,144 & 377.4 \\
 & Mixed topics & 5.6\% & 24,309 & 56.1 \\
 & Podcast & 2.7\% & 9,790 & 27 \\
 & Review & 5.7\% & 26,917 & 56.9 \\
 & TV show & 1.7\% & 5,877 & 16.9 \\
\midrule
\multirow{2}{*}{\rotatebox[origin=c]{90}{\scriptsize Lang.}}
 & Vietnamese & 52.9\% & 239,476 & 530.9 \\
 & Code-switching & 47.1\% & 170,290 & 472.7 \\
\midrule
\multirow{2}{*}{\rotatebox[origin=c]{90}{\scriptsize Gender}}
 & Male & 58.4\% & 246,066 & 576.9 \\
 & Female & 41.6\% & 157,875 & 416.5 \\
\bottomrule
\end{tabular}
\caption{\datasetname{} distribution by dialect, topic, language, and gender.
Percentages are duration-based except gender, which is speaker-based. Code-switched segments average 2.83 English words (8.12\% of words).}
\label{tab:dataset-distributions}
\end{table}

\begin{table}[tb!]
\centering
\footnotesize
\begin{tabular}{@{}lccrrrr@{}}
\toprule
Synthesizer & MOS $\uparrow$ & Spk. Sim. $\uparrow$
& Videos & Hours & \#Spk. & \#Seg. \\
\midrule
Bona fide & 2.88 & --- & 8,388 & 993.4 & 1,262 & 403,941 \\
MiniMax & 2.93 & 0.55 & 8,388 &  1,032.9 & 1,262 & 403,941 \\
OmniVoice & 2.54 & 0.74 & 4,610 & 718.7 & 810 & 293,358 \\
Higgs Audio v3 & 2.85 & 0.72 & 4,610 & 719.26 & 810 & 293,881 \\
VoxCPM2 & 2.42 & 0.76 & 4,609 & 719.26 & 810 & 293,878 \\
\midrule
Overall (spoof) & 2.71 & 0.68 & 8,388 & 3,190 & 1,262 & 1,290,883 \\
\bottomrule
\end{tabular}%
\caption{Naturalness (via MOS), speaker similarity (Spk. Sim.), and scale before
similarity filtering. MOS uses a five-point scale for naturalness. Overall scores are
segment-weighted, with deduplicated video, speaker counts (\#Spk) and segment counts (\#Seg.).}
\label{tab:full-dataset-summary-fake}
\vspace{-5pt}
\end{table}

\subsection{Statistical Summary and Uniqueness}

Table~\ref{tab:dataset-distributions} summarizes \datasetname{}'s statistics. 
Among our 1,262 verified speakers, 568 have North dialect, 547
have South dialect, 78 have Central dialect, 30 have Southwest dialect,
and 39 have Overseas dialect. The North and South groups provide broad
coverage of the two largest varieties, while other identified groups enable evaluation beyond the dominant dialects. Noticeably, to the best of our knowledge, \datasetname{} \textit{is the only
listed resource combining released transcripts, natural code-switching, and
explicit dialect coverage.} Its spoof subset is approximately nine times larger
by utterances and eighteen times larger by duration than the next-largest
released Vietnamese subset, while uniquely providing \textit{speaker-matched and transcript-match} bona fide--spoof pairs from four synthesizers (Table~\ref{tab:unified-datasets-compact}). We further report the spoof synthesis quality in Table~\ref{tab:full-dataset-summary-fake}, demonstrating both scale and diversity of \datasetname{} in enabling practical deepfake benchmark for Vietnamese.

Compared with existing datasets, \datasetname{} provides
substantially greater coverage of the three major dialects.

For instance, \datasetname{} provides 502.5, 467.3, and 20.0 hours,
respectively, or about 5.0, 5.9, and 3.0 times as much compared to Vietnam-Celeb's
\cite{pham2023vietnam}. Although these categories are not directly comparable, \datasetname{} additionally
provides 7.3 hours of Southwest and 6.5 hours of Overseas Vietnamese speech,
with the latter explicitly annotated as a separate group, enhancing the diversity of the dataset.

\begin{table}[tb!]
\centering
\footnotesize
\setlength{\tabcolsep}{3pt}
\begin{tabular}{@{}l cc c ccc c@{}}
\toprule
\multirow{2}{*}{\textbf{Generator}} & \multicolumn{2}{c}{\textbf{DFA}} && \multicolumn{3}{c}{\textbf{ADF}} & \multirow{2}{*}{\textbf{Mean}} \\
\cmidrule(lr){2-3} \cmidrule(lr){5-7}
 & \textbf{1B} & \textbf{500M} && \textbf{W2V2-L} & \textbf{XLS-R-2B} & \textbf{MMS-300M} & \\
\midrule
MiniMax   & 13.14 & 16.60 && 47.12 & 56.30 & 52.13 & 37.06 \\
OmniVoice & 10.95 & 15.10 && 0.01  & 0.00  & 50.84 & 15.38 \\
HIGGS     & 12.74 & 16.48 && 79.97 & 0.02  & 1.35  & 22.11 \\
VoxCPM2   & 34.33 & 38.52 && 0.00  & 0.00  & 99.99 & 34.57 \\
\midrule
Mean      & 17.79 & 21.67 && 31.77 & 14.08 & 51.08 & 27.28 \\
\bottomrule
\end{tabular}
\caption{Zero-shot detector performance in \% EER $\downarrow$.}
\label{tab:zero-shot-by-generator}
\end{table}

\section{Experiments}
\label{sec:experiments}

\noindent \textbf{Evaluation of Multilingual Deepfake Detectors.} We benchmark five pretrained multilingual detectors in zero-shot setting without fine-tuning, all of which was trained with Vietnamese: DF-Arena (DFA)
1B and 500M~\cite{kulkarni2026compactsslbackbonesmatter}, and three
AntiDeepfake (ADF) checkpoints~\cite{ge2025post}. Each detector is
evaluated against our synthesizers using matched bona fide--spoof pairs. Table~\ref{tab:zero-shot-by-generator} summarizes the results. Overall, DFA-1B is the most balanced across generators, with 10.95--13.14\% EER on
three synthesizers but 34.33\% on VoxCPM2. Similarly, ADF-XLS-R-2B has the best mean EER and accuracy, yet ranges from chance performance against MiniMax to perfect accuracy on
OmniVoice and VoxCPM2, indicating \textit{generator-specific rather
than uniform generalization.}

\begin{figure}[t]
\centering
\includegraphics[width=0.9\columnwidth]{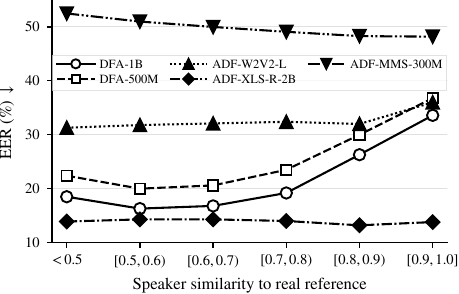}
\caption{Zero-shot EER by speaker similarity between synthesized speech and real reference, averaged across synthesizers. 
}
\label{tab:zero-shot-similarity-stratification}
\end{figure}

\vspace{3pt}
\noindent \textbf{Similarity-Stratified Evaluation.} Spoofs with similarity scores are divided into six disjoint ranges. Fig~\ref{tab:zero-shot-similarity-stratification} shows that higher similarity degrades several detectors: from $[0.5,0.6)$ to $[0.9,1.0]$,
DFA-1B EER rises from 16.3\% to 33.6\% with DFA-500M showing similar trend.
However, ADF-family detectors show to be much more stable. Regardless, compared with state-of-the-art deepfake detection results in languages such as English, our reported EERs are much inferior, revealing the practical utility of our collected dataset in benchmarking and advancing spoof speech detection for Vietnamese.

\vspace{3pt}
\noindent \textbf{Dialect-Stratified Evaluation.} 
Table~\ref{tab:cross-dialect-generalization} 
We observe that, overall, the patterns differ across dialects with several detectors weaker on Southwest and Overseas Vietnamese: DFA-1B EER
rises to 28.7\% and 26.9\%, versus 16.6\% on South and 18.3\% on North, with
similar trends for DFA-500M. However, Central, which is less popular than South and North dialect, performs comparably to larger dialect subsets for several models, while some ADF models
remain stable or improve on rarer dialects. Thus, dialect effects are
model-dependent and may reflect recording conditions, speaker composition,
synthesis coverage, and acoustics rather than corpus frequency alone.

\begin{table}[t]
\centering
\footnotesize
\setlength{\tabcolsep}{1.3pt}
\begin{tabular}{@{}lcccccc@{}}
\toprule
Detector & South & Southwest & North & Central &
\begin{tabular}[c]{@{}c@{}}Overseas\\Vietnamese\end{tabular} & Macro \\
\midrule
\# Bona fide & 173,195 & 4,431 & 215,266 & 8,877 & 2,172 & -- \\
\midrule
ADF-XLS-R-2B  & 13.7 & 14.4 & 14.4 & 13.1 & 13.4 & \textcolor{blue}{\textbf{13.8}} \\
DFA-1B        & 16.6 & 28.7 & 18.3 & 17.5 & 26.9 & 21.6 \\
DFA-500M      & 21.0 & 33.2 & 21.7 & 22.3 & 28.3 & 25.3 \\
ADF-W2V2-L    & 31.2 & 35.5 & 32.0 & 33.3 & 14.2 & 29.2 \\
ADF-MMS-300M  & 52.3 & 50.5 & 50.4 & 49.7 & 48.8 & 50.3 \\
\midrule
Average       & 26.96 & \textcolor{red}{\textbf{32.46}} & 27.36 & 27.18 & \textcolor{blue}{\textbf{26.32}} & 28.04 \\
\bottomrule
\end{tabular}%
\caption{Zero-shot EER (\%, $\downarrow$) by dialect, averaged
equally across generators. Macro is the unweighted dialect average.
Detectors are sorted by Macro EER. Red and blue denotes the lowest and highest EER in respective column and row.}
\label{tab:cross-dialect-generalization}
\end{table}

\begin{figure}[tb!]
\centering
\includegraphics[width=\columnwidth]{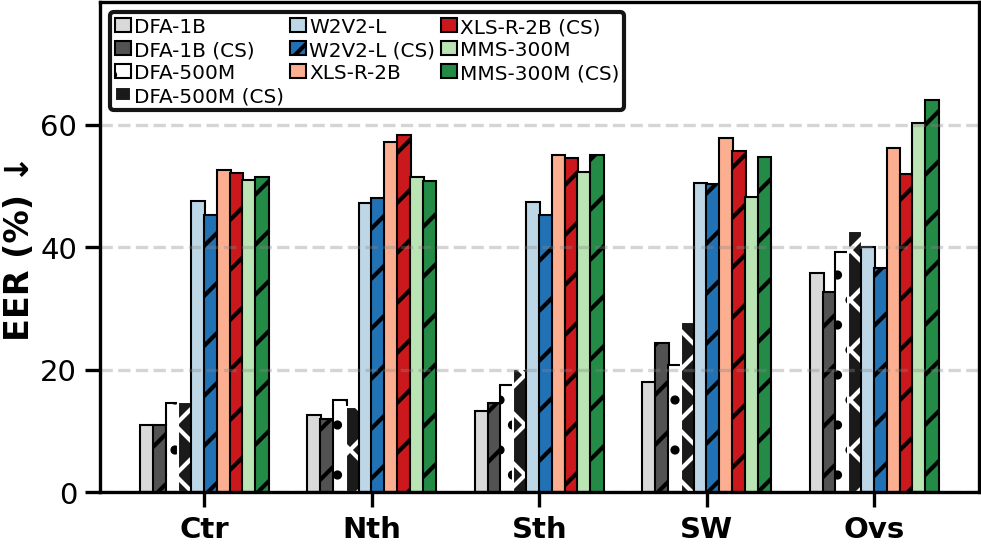}
\caption{DFA's EER (\%) in zero-shot detection of MiniMax deepfake on code-switching (CS) utterances across dialects.}
\label{tab:zero-shot-code-switch-stratification}
\end{figure}

\vspace{3pt}
\noindent \textbf{Performance with Code-Switching.} We observe that MiniMax, the only commercial synthesizer that we test, exhibits a unique resistance to detection under code-switching conditions compared against open-source models. Across dominant regional dialects like South and Southwest Vietnamese, code-switching consistently increases MiniMax's EER (e.g., up to +7.1\% points on DFA-500M) (Fig.~\ref{tab:zero-shot-code-switch-stratification}). This suggests that proprietary commercial architectures likely achieve better phoneme blending and acoustic smoothing during language switches, successfully hiding synthetic cues that open-source models inadvertently reveal.

\section{Other Potential Use Cases}
\label{sec:other-applications}

Beyond deepfake detection, \datasetname{} supports ASR, language modeling,
speaker verification, diarization, and speaker-disjoint deepfake evaluation or speaker verification for Vietnamese. Its dialect, code-switching also enable accent modeling, voice-cloning evaluation, synthesis attribution, and fairness studies. Our unique speaker-matched bona fide--spoof annotations can potentially help isolate and reduce linguistic sensitivity of deepfake detectors trained with no difference in utterances, rather than those that often trained with real and spoof utterances of very different distributions.

\section{Conclusion}
\label{sec:conclusion}

We presented \datasetname, a multi-domain Vietnamese corpus containing 993.4
hours and 403,941 bona fide segments from 1,262 speakers, with transcripts,
dialect and speaker metadata, and natural code-switching. It also provides
2,765.74 hours of speaker-matched spoof speech from four synthesis systems.
Results show that detector performance depends on the generator, dialect, and
speaker similarity, motivating evaluation across diverse conditions.

\bibliographystyle{IEEEbib}
\bibliography{refs}

\end{document}